\documentclass[preprint,12pt]{elsarticle}

\usepackage{amssymb}
\usepackage{amsmath}
\usepackage{float}

\begin{document}

\begin{frontmatter}



\title{A Hybrid Real-Time Framework for Efficient Fussell-Vesely Importance Evaluation Using Virtual Fault Trees and Graph Neural Networks}


\author[1]{Xingyu Xiao} 

\affiliation[1]{organization={Institute of Nuclear and New Energy Technology,Tsinghua University},
            city={Beijing},
            postcode={100084}, 
            country={China}}

\author[2]{Peng Chen} 

\affiliation[2]{organization={Software Institute of Chinese Academy of Sciences},
            city={Beijing},
            postcode={100086}, 
            state={Beijing},
            country={China}}

\begin{abstract}
The Fussell-Vesely Importance (FV) reflects the potential impact of a basic event on system failure, and is crucial for ensuring system reliability. However, traditional methods for calculating FV importance are complex and time-consuming, requiring the construction of fault trees and the calculation of minimal cut set. To address these limitations, this study proposes a hybrid real-time framework to evaluate the FV importance of basic events. Our framework combines expert knowledge with a data-driven model. First, we use Interpretive Structural Modeling (ISM) to build a virtual fault tree that captures the relationships between basic events. Unlike traditional fault trees, which include intermediate events, our virtual fault tree consists solely of basic events, reducing its complexity and space requirements. Additionally, our virtual fault tree considers the dependencies between basic events rather than assuming their independence, as is typically done in traditional fault trees. We then feed both the event relationships and relevant data into a graph neural network (GNN). This approach enables a rapid, data-driven calculation of FV importance, significantly reducing processing time and quickly identifying critical events, thus providing robust decision support for risk control. Results demonstrate that our model performs well in terms of MSE, RMSE, MAE, and R², reducing computational energy consumption and offering real-time, risk-informed decision support for complex systems.

\end{abstract}

\begin{keyword}


Fussell-Vesely Importance  \sep Minimal Cut Sets \sep Interpretive Structural Modeling \sep Graph Neural Networks \sep Reliability Engineering
\end{keyword}

\end{frontmatter}

\section{Introduction}
\label{Introduction}

Ensuring the reliability of complex systems is a critical concern across numerous industries, including aerospace, energy, manufacturing, transportation and nuclear power plants (NPPs). System failures in these contexts can have severe consequences, leading to not only financial losses but also endangering human lives and causing environmental damage \cite{dewey2008}. As a result, reliability analysis has emerged as a vital discipline aimed at identifying and mitigating potential risks. The Fussell-Vesely (FV) importance \cite{tyrvainen2013} is one of the most widely used indicators for evaluating the significance of these basic events in the context of overall system reliability . Specifically, FV importance quantifies the contribution of a single basic event to the probability of system failure, making it an invaluable tool for risk prioritization and decision-making \cite{budiman2022}.

Despite significant advancements in system reliability analysis and risk management, there remain notable gaps in current methodologies, particularly in the calculation of Fussell-Vesely (FV) importance for complex systems. Traditional reliability approaches, such as fault tree analysis (FTA) \cite{ruijters2015}, while widely used, face several limitations. Chief among these challenges is the computational complexity and time-intensive nature of calculating FV importance in large, interconnected systems. Traditional methods often require the construction of detailed fault trees \cite{klamt2020}, the identification of minimal cut sets \cite{vatn1992}, and the assumption of event independence \cite{clarotti1981}. These assumptions and requirements limit the applicability of traditional approaches in modern, dynamic environments where real-time risk assessment \cite{backstrom2016} is crucial.

One of the primary gaps identified in the literature is the reliance on either expert-driven models, such as Interpretive Structural Modeling (ISM), or purely data-driven techniques, such as machine learning, without fully leveraging the strengths of both. ISM has proven effective for simplifying complex systems by capturing expert knowledge and organizing relationships between events in a hierarchical structure. However, ISM alone does not provide the data-driven adaptability required for real-time system monitoring and dynamic FV importance evaluation. While ISM effectively models dependencies and structures, it lacks the capability for rapid recalculation and updating based on incoming data streams, which is necessary for real-time applications.

On the other hand, data-driven models, particularly machine learning techniques like Graph Neural Networks (GNN), have shown promise in processing graph-structured data and capturing the interdependencies among system components. GNNs are particularly well-suited for reliability analysis because they can handle dynamic graphs and continuously learn from new data. However, while GNNs excel at pattern recognition and prediction, they often operate as black-box models, lacking transparency and interpretability \cite{grimstad2024}. Furthermore, data-driven approaches may struggle in domains where historical data is scarce or incomplete, thus relying solely on such models may lead to unreliable or biased predictions.

Thus, a clear research gap exists in the development of hybrid frameworks that can combine the structural clarity of expert knowledge with the adaptability and real-time capabilities of data-driven models. 
In response to these challenges, recent advancements in data-driven approaches and machine learning models offer promising alternatives for improving efficiency and accuracy in reliability assessment. This study proposes a novel hybrid real-time framework designed to evaluate the FV importance of basic events in complex systems. By integrating expert knowledge through Interpretive Structural Modeling (ISM) and leveraging the data-driven capabilities of a Graph Neural Network (GNN), our framework circumvents the limitations of traditional fault tree analysis. Unlike conventional approaches, our virtual fault tree focuses solely on basic events and captures their interdependencies, allowing for a more streamlined and flexible assessment of system reliability.

This new approach has the potential to revolutionize the process of calculating FV importance by significantly reducing computational complexity and time. Moreover, by incorporating real-time data and expert insights, our method provides more accurate and timely decision support for risk control in dynamic environments. This study evaluates the performance of our model across multiple metrics, including Mean Squared Error (MSE), Root Mean Squared Error (RMSE), Mean Absolute Error (MAE), and R², demonstrating its effectiveness in reducing computational energy consumption while maintaining high levels of accuracy. By bridging the gap between expert-driven and data-driven methods, this framework enables rapid, accurate, and adaptive reliability assessments, providing real-time decision support for risk management in dynamic and complex systems..

The remainder of this paper is organized as follows: Section \ref{Literature Review} introduces the literature review. Section \ref{Methodology} details the hybrid framework combining ISM and GNN for real-time FV importance evaluation. Section \ref{Experimental Setup} outlines the data and methods used to evaluate the performance of the proposed framework. Section \ref{Results and Analysis} presents the findings from the experiments conducted using the proposed framework. Section \ref{Conclusion and Discussion} concludes by summarizing the key contributions.

\section{Literature Review}
\label{Literature Review}

\subsection{ Fussell-Vesely Importance Measures in Fault Trees}

Fussell-Vesely (FV) importance \cite{wang2013fuzzy} is the most commonly used importance measures in fault trees, as it directly reflects the contribution of basic events to system failure and is closely associated with minimal cut sets \cite{test}. Specifically, it is a quantitative metric used to evaluate the extent to which a basic event (or component) contributes to the top event (system failure) in a system. It represents the proportion of the total probability of the top event that is attributable to the occurrence of a specific basic event.

The methods for calculating FV importance can generally be categorized into three types: analytical methods, structural modification algorithms, and statistical simulation algorithms. Analytical methods include Fault Tree Analysis (FTA) and minimal cut sets (MCS) \cite{fussell1975}, both are suitable for small-scale systems and single failure models. However, as system complexity increases, the computational complexity grows exponentially, making it difficult to scale these methods to large and complex systems.

Structural modification algorithms optimize the entire computational process by modifying the structure of the fault tree. These include Binary Decision Diagrams (BDD) \cite{jung2004}, Bayesian Networks (BN) \cite{bobbio2001}, and sparse matrix methods \cite{parter2013}, which map the fault tree to BDD, BN, and sparse matrices, respectively, to improve computational efficiency and the dynamic characteristics of the algorithms. However, the process of converting to BDD is rather complex \cite{reay2002}, and the BN approach has high computational complexity, with a complicated model design and inference process \cite{khakzad2011}. Sparse Matrix Methods are suitable for sparse structures but are not well-suited for dense fault trees \cite{parter2013}.

Statistical simulation algorithms calculate by sampling simulations, including monte carlo simulations \cite{rao2009}, importance sampling \cite{zhao2024}, Markov Chain Monte Carlo (MCMC) \cite{yevkin2016}, and heuristic algorithms \cite{mo2013}. This approach offers high flexibility and is applicable to various system structures; however, it often suffers from long convergence times, low sampling efficiency in complex systems, and high computational costs due to the large number of simulations required.

Therefore, an efficient method capable of calculating FV importance for complex systems is required. This study proposed a hybrid real-time framework for efficient FV importance evaluation using virtual fault trees and graph neural networks. This method overcomes the limitations of traditional approaches in terms of complexity, time efficiency, and accuracy, offering a new research direction for future reliability analysis.

\subsection{Interpretive Structural Modeling and Expert Knowledge}

Interpretive Structural Modeling (ISM) has been widely adopted in various fields for its ability to simplify the analysis of complex systems by breaking them down into manageable, hierarchical structures. Several studies have demonstrated the effectiveness of ISM in diverse domains, such as supply chain management, risk assessment, and system engineering. For instance, Deng et al. \cite{deng2023national} employed ISM to model the assessment of the COVID‐19 pandemic. Similarly, Ma et al. \cite{ma2024international} used ISM to quantify international relations. 

In the field of system reliability, ISM has been used to simplify the analysis of large, interconnected systems. Traditional methods like fault tree analysis can become unmanageable when the number of components and their interactions grow exponentially. ISM helps mitigate this complexity by reducing the system into a structured model, focusing on the relationships between basic events and eliminating the need to model intermediate events explicitly. This reduction in complexity makes ISM a valuable tool in enhancing the clarity and efficiency of system modeling.

Moreover, previous studies have highlighted the importance of expert input in reliability modeling and risk assessment. For example, Xiao et al. \cite{xiao2024emergency} demonstrated how expert knowledge was used to assess the reliability of complex electrical systems, where the lack of extensive failure data required an expert-driven analysis of potential risks. In ISM, expert knowledge is integral to constructing accurate structural models. Therefore, we employed ISM to construct a virtual fault tree that includes only the relationships between basic events. This approach reduces the consumption of spatial resources and takes into account the correlations between basic events. Additionally, the construction of the virtual fault tree is expert-friendly.

\subsection{Graph Neural Networks (GNN) for System Reliability}

Recent advancements in machine learning have resulted in the creation of models tailored to process complex, non-Euclidean data structures, notably graphs. Among these, Graph Neural Networks (GNNs)\cite{li2023graph} have attracted significant attention for their capacity to effectively represent and analyze data structured as graphs. Unlike traditional neural networks that process grid-like data, such as images or sequences, GNNs are particularly adept at handling graph data, which inherently captures complex relationships between elements. This renders GNNs highly suitable for a broad spectrum of applications that hinge on the intricate interactions between components or events \cite{zhou2020graph}.

In the context of system reliability, GNNs provide a powerful framework for modeling and analyzing the intricate interdependencies between system components. Graph Neural Networks have already found success in various domains where reliability and risk management are crucial, such as fault diagnosis in power systems \cite{liu2024causal, nguyen2023spatial}, recommender systems \cite{wu2022graph}.  GNNs can be trained on temporal graph data, allowing for real-time updates to reliability models as new information becomes available. This adaptability makes GNNs particularly advantageous for real-time risk management and reliability assessments.

In system reliability, the application of GNNs to fault tree analysis is an emerging area of research. Traditional fault trees assume independent failures between basic events, but in many systems, components exhibit dependencies that cannot be ignored. GNNs, with their inherent ability to model dependencies between nodes, offer a powerful alternative by analyzing the entire fault tree as a graph. This allows for more accurate predictions of system failure probabilities by accounting for the interconnectedness of events and their impact on overall system reliability \cite{grimstad2024}.

Overall, GNNs provide a promising solution for modeling complex systems where the relationships between components play a critical role in system reliability. By leveraging their ability to learn from graph-structured data, GNNs offer significant advantages over traditional methods in terms of flexibility, scalability, and accuracy, making them a key tool for improving risk management and ensuring system reliability across various industries.

\section{Methodology}
\label{Methodology}

\subsection{Overview of the Proposed Framework}

Our framework integrates expert knowledge with a data-driven model. First, we leverage expert knowledge to construct a virtual fault tree using the ISM model, which represents the relationships between basic events. Unlike traditional fault trees that include intermediate events, our virtual fault tree only contains basic events, thereby occupying less space. Furthermore, our virtual fault tree takes into account the relationships between basic events, rather than assuming that these events are independent as in traditional fault trees. Next, we input the relationships between the basic events, along with the data, into a deep learning network—specifically, a graph neural network. This enables the rapid, data-driven calculation of FV importance, reducing the time required for the entire process and quickly identifying critical events, ultimately providing decision support for risk control.

\begin{figure}[H]
\centering
\includegraphics[width=1.0 \textwidth]{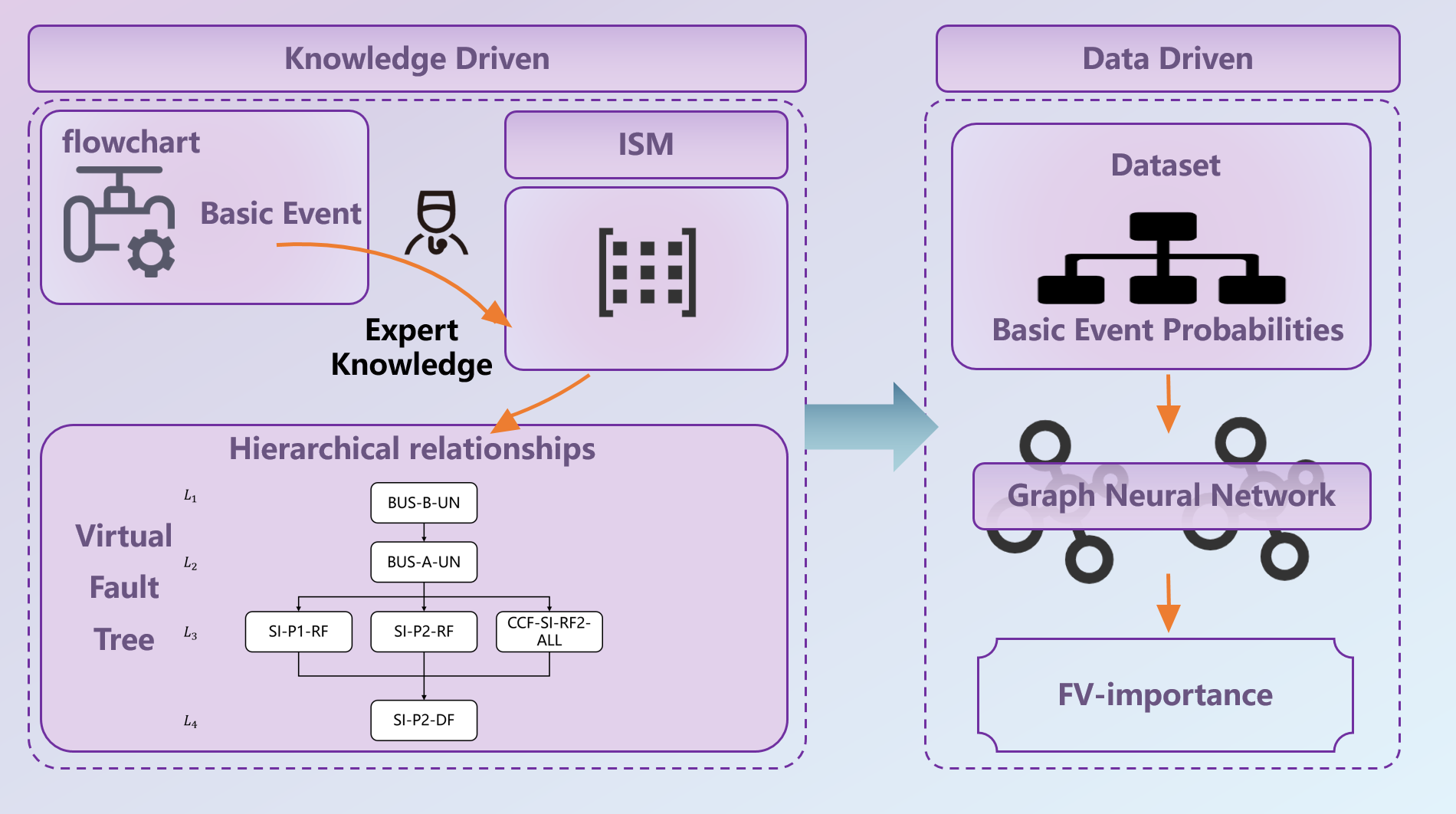}
\caption{Hybrid framework for real-time fussell-vesely importance evaluation integrating interpretive structural modeling and graph neural networks}\label{levels}
\end{figure}

\subsection{ISM-Based Virtual Fault Tree Construction}
The first step in our framework involves leveraging expert knowledge to build a virtual fault tree using Interpretive Structural Modeling (ISM) \cite{singh2008knowledge} . ISM is a well-established methodology that systematically captures and organizes the relationships between variables—in this case, basic events—through expert input. Unlike traditional fault trees that include intermediate events, our virtual fault tree focuses exclusively on basic events, reducing both the complexity and space requirements of the model.

The ISM method is often used for hierarchical structure analysis of complex systems, making it particularly suitable for systems involving multiple factors and layers. Specific application areas include risk management \cite{delahuerga2015, song2020ism}, project management \cite{mannan2013knowledge, shakeri2020analysis}, decision support \cite{mishra2020integrated, vinodh2021development}, and others. It is an expert knowledge-driven approach. It serves as a modeling tool for analyzing the structure of complex systems. This method helps to understand the system's structure and hierarchy by establishing the relationships between elements and representing these relationships in a graphical form. 

The ISM process begins by identifying the key elements within a system, followed by establishing the relationships between them through expert judgment, literature review, or data analysis. These relationships are represented in an adjacency matrix, which is subsequently transformed into a reachability matrix through mathematical operations to reveal both direct and indirect connections. Next, hierarchical partitioning is performed based on the reachability matrix, organizing the elements into levels according to their influence. Finally, the system's elements and their interactions are illustrated using a directed acyclic graph (DAG) to depict the hierarchical structure and relationships.

Using ISM, we generate a directed graph that illustrates the structural relationships among the basic events. This graph reflects the hierarchical dependencies between events and incorporates any causal or sequential relationships that could affect system reliability. ISM's ability to visualize and formalize these interactions between events provides a robust foundation for subsequent data-driven analysis while ensuring that expert insights are captured early in the process. Additionally, ISM enables us to account for the inherent dependencies between events, which is a significant departure from the independence assumptions made in traditional fault tree analysis.

\subsection{Graph Neural Network for FV Importance Evaluation}

At the same time, our framework also incorporates data-driven methods. Once the virtual fault tree is constructed using ISM, the next step is to transition from expert knowledge to a data-driven approach. Graph Neural Networks (GNN) are a deep learning method designed for handling graph-structured data, making them well-suited for our virtual fault tree analysis. We feed the graph structure, along with relevant system data, into a Graph Neural Network (GNN). GNNs are a class of machine learning models specifically designed to process data in graph format, making them highly suitable for analyzing the interdependencies and relationships between basic events in our virtual fault tree. The core idea of GNNs is to aggregate the features of each node with the features of its surrounding nodes to form a new node representation. This process can be achieved through message passing, where each node receives messages from its neighboring nodes and aggregates these messages into a new node representation. This method can be iterated multiple times to capture more comprehensive information from the graph structure.

We chose GCN (Graph Convolutional Networks) for learning the F-V importance. GCN is a spectral domain-based graph convolution method. It utilizes the Laplacian matrix of the graph, performing convolution operations on the feature vectors to enable information transfer and feature extraction. 

In GCN, the feature vector of each node is weighted and averaged with the feature vectors of its neighboring nodes. The weights are determined by the values in the adjacency matrix A. Specifically, the convolution operation in GCN can be expressed in Equation \ref{GCN}.

\begin{equation}
Z = f(A,X) = D^{-1} A X W
\label{GCN}
\end{equation}

Where, $D$ is the degree matrix of $A$, and $W$ is the learnable weight matrix.

GCN has been widely applied to tasks such as node classification \cite{hu2019hierarchical,abu2020n}, graph classification \cite{abu2020n,yao2019graph}, and link prediction \cite{lei2019gcn,li2019mv}, achieving excellent results.

\subsection{Real-Time Calculation and Decision Support}
The primary advantage of integrating ISM and GNN into a hybrid framework is the ability to perform real-time evaluations of FV importance. As the GNN is continually updated with new data, it dynamically adjusts the importance rankings of basic events, offering real-time insights into the system’s risk profile. This capability is particularly valuable for dynamic systems where operating conditions and risk factors are subject to frequent changes. 
Moreover, by streamlining the calculation process, our framework significantly reduces the computational overhead associated with traditional FV methods. The elimination of intermediate events in the fault tree and the reliance on a GNN model ensure that our approach can rapidly evaluate the contribution of individual events to system failure. This efficiency enables real-time decision support for risk mitigation, allowing system operators to prioritize maintenance, inspections, or other preventive measures based on the current reliability status of key events.

\subsection{Performance Evaluation and Validation}
\label{Performance Evaluation and Validation}

To validate the performance of our model, MSE (Mean Squared Error), RMSE (Root Mean Squared Error), MAE (Mean Absolute Error), and $R^{2}$  (R-squared, coefficient of determination) are introduced in this paper. They are commonly used evaluation metrics in machine learning and statistical models for regression. They are used to measure the prediction error and goodness of fit of a model. Additionally, we assess the computational efficiency of our approach by evaluating its processing time and energy consumption in comparison to fault tree analysis and other traditional techniques.

To ensure the effectiveness of the proposed framework, we conduct a thorough performance evaluation. We measure the accuracy of the FV importance calculations by comparing the framework's outputs against traditional methods, using metrics such as Mean Squared Error (MSE), Root Mean Squared Error (RMSE), Mean Absolute Error (MAE), and R². Additionally, we assess the computational efficiency of our approach by evaluating its processing time and energy consumption in comparison to fault tree analysis and other traditional techniques.
Results from our experiments demonstrate that the hybrid framework not only achieves high accuracy in FV importance prediction but also significantly reduces computational time and energy consumption. Furthermore, the dynamic nature of our approach offers continuous, risk-informed decision support, enabling more effective and timely risk management in complex systems.

\section{Experimental Setup}
\label{Experimental Setup}

To validate the applicability of our framework, we conducted experimental testing on a high-risk system in this section. Specifically, we analyzed a simplified system diagram for a nuclear power plant.

\subsection{System Description}

\begin{figure}[H]
\centering
\includegraphics[width=0.8\textwidth]{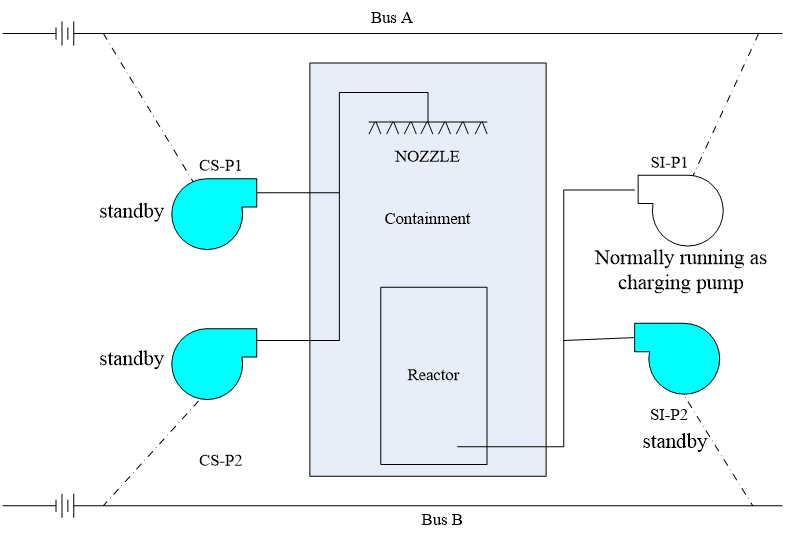}
\caption{ Simplified system diagram for a nuclear power plant}\label{diagram}
\end{figure}

The diagram of the system is shown in Figure \ref{diagram}. It includes two subsystems: the safety injection (SI) system and the containment spray (CS) system. The SI system is equipped with two identical pumps, designated as SI-P1 and SI-P2. Under normal conditions, SI-P1 operates as a charging pump while SI-P2 remains in standby. In the event of a SI condition, both pumps function as SI pumps. The system has two potential failure modes for each pump: a failure to start, with a probability of 5E-5 per demand, and a failure while running, with a probability of 1E-6 per hour. Unavailabilities resulting from maintenance, testing, or human error are excluded from consideration. However, common cause failure between the pumps must be accounted for to ensure the system's reliability.

Additionally, Bus A and Bus B have a failure mode characterized by failure during operation, with a probability of 1E-7 per hour. During normal operation, either Bus A or Bus B can be taken out for maintenance, though both cannot be maintained simultaneously. The average maintenance interval is every three months, and the average maintenance duration is approximately 20 minutes. Common cause failures between Bus A and Bus B are excluded from consideration.

\subsection{ISM-Based Virtual Fault Tree Construction}

Consequently, we have systematically categorized the failure modes of the aforementioned system, as shown in Table \ref{failure_mode}. The failure modes are classified into two types: fail to start and fail during operation. The designations represent the failure states of different equipment under various failure modes. 

\begin{table}[H]
\centering 
\begin{tabular}{lrll}
\hline
Index & \multicolumn{1}{l}{Failure Mode}  & Equipment   & Designation    \\
\hline
1     & \multicolumn{1}{l}{Fail to start} & SI-P2       & SI-P2-DF       \\
\cline{1-2} 
2     &     & SI-P1       & SI-P1-RF       \\
3     &                                   & SI-P1       & SI-P2-RF       \\
4     &    Fail running          & SI-P1 \& SI-P2 & CCF-SI-RF2-ALL \\
5     &       & BUS A       & BUS-A-UN       \\
6     &     & BUS B       & BUS-B-UN  \\
\hline
\end{tabular} \caption{Designations for different equipment with different failure modes}\label{failure_mode}
\end{table}

1. Construct a fault tree for the system

The fault tree constructed for the SI system is shown in Figure \ref{ET}. Since SI-P1 can normally operate as a charging pump, the fail-to-start failure mode was not considered for SI-P1; only the fail-to-run failure mode was taken into account. For SI-P2, two failure modes were considered, as detailed in Table \ref{failure_mode}. The fail-to-run failures of Bus A and Bus B and their impacts on both SI-P1 and SI-P2 were also incorporated into this analysis. Additionally, the common cause failure of SI-P1 and SI-P2, denoted as SI-RF2-ALL, was included, as indicated by the triangle in the figure.

\begin{figure}[H]
\centering
\includegraphics[width=0.8 \textwidth]{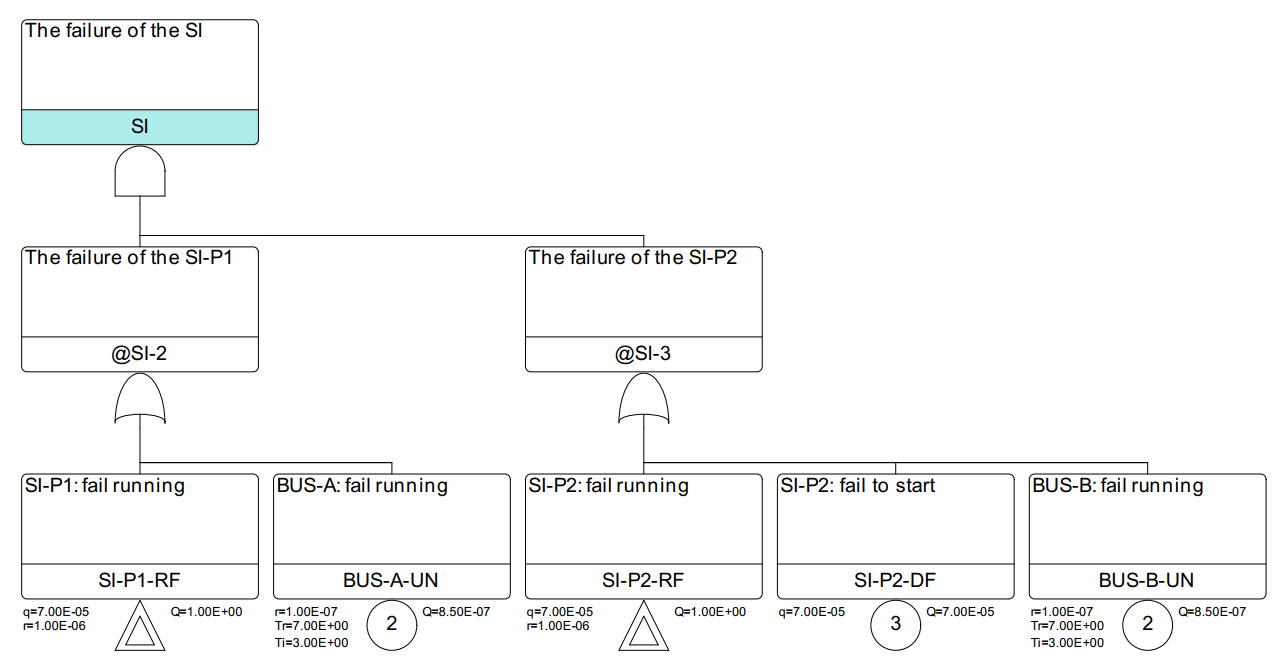}
\caption{	Fault tree for the safety injection system in the case}\label{ET}
\end{figure}

\begin{figure}[H]
\centering
\includegraphics[width=0.8 \textwidth]{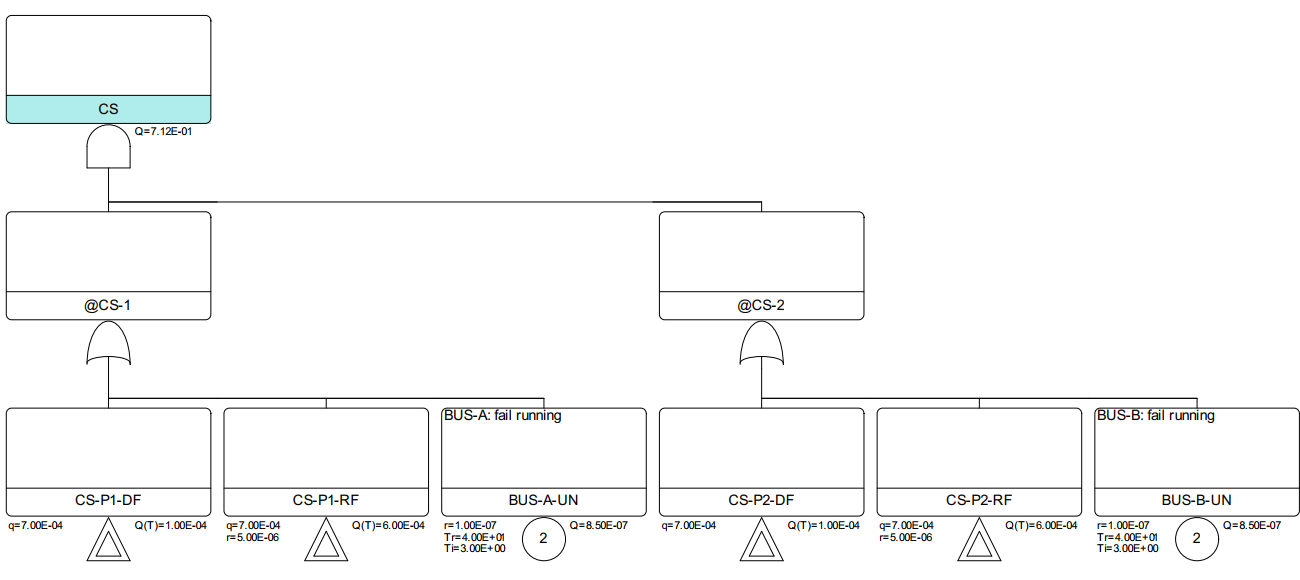}
\caption{	Fault tree for the containment spray system in the case}\label{ET}
\end{figure}

2. Data collection.

We utilized Risk Spectrum (Version 1.5.4) to simulate the Fussell-Vesely (F-V) importance of basic events under different probabilities. It is known that in the Risk Spectrum, the types of reliability parameters include probability \cite{bertsekas2008introduction}, failure rate \cite{finkelstein2008failure}, frequency \cite{van1927frequency}, MTTR \cite{kullstam1981availability}, test interval \cite{paylor2006use}, and others. Among these, probability represents the constant unavailability, such as demand failure rate; failure rate represents the failure rate (1/h); frequency represents the occurrence frequency of initiating events (1/year); MTTR represents the mean time to repair (h); and test interval represents the testing interval (h).

The distribution types considered include lognormal, Gamma, etc. In this experiments, only the lognormal distribution was used. The expression for the lognormal distribution is shown in Equation\ref{lognormal}.

\begin{equation}
f(y) = \frac{1}{\sigma y\sqrt{2\pi } } exp\left[ -\frac{1}{2\sigma^{2 } } (lny-\mu )  \right ] 
\label{lognormal}
\end{equation}

where, $\mu$ represents the mean value, while $\sigma$ is the error factor.

The initial parameter values for each basic event are provided in the Table~\ref{parameter}. Data was generated by fluctuating these values within the same order of magnitude. We first utilized RiskSpectrum PSA Version 1.5.4 to input the fundamental information. We then simulated the FV importance measures under varying probabilities of basic event occurrence. Following this, we employed the software to generate minimal cut sets under different conditions. For the CS system, we collected 304 data points, and for the SI system, we collected 316 data points.

It is important to mention that the RiskSpectrum PSA software is proprietary. As a result, we were unable to automate the process and had to manually adjust the parameters to generate the data. This manual process was time-consuming, taking nearly a week to complete the data collection.

\begin{table}[H] 
\centering 
\begin{tabular}{p{1cm} p{4cm} p{2.5 cm} p{2cm} p{2cm}} 
\hline
Item & Parameter & Calculation Type & Mean Value & Distribution Type \\ 
\hline
1 & QQ-CS-Pump & Probability & 1.00E-04 & Lognormal \\
2 & QQ-SI-pump & Probability & 5.00E-05& Lognormal  \\
3 & RR-bus  & Failure Rate & 1.00E-07& Lognormal  \\
4 & RR-CS-Pump & Failure Rate & 5.00E-06 & Lognormal\\
5 & RR-SI-Pump& Failure Rate & 1.00E-06 & Lognormal \\
6 & FF-LLOCA & Frequency& 1.00E-04 & Lognormal\\
7 & TR-BUS & MTTR &2.00E+01& Lognormal \\
8 & TI-BUS & Test Interval & 3.00E+00 & Lognormal\\
9 & BETA-CS-PUMP &  Beta Factor & 1.00E-01 & Lognormal\\
10 & BETA-SI-PUMP & Beta Factor  & 5.00E-02 & Lognormal\\
\hline
\end{tabular}
\caption{The initial information for all basic events in the CS and SI systems}
\label{parameter}
\end{table}

3. Established the relationships among the basic events based on ISM

\begin{figure}[H]
\centering
\includegraphics[width=0.6 \textwidth]{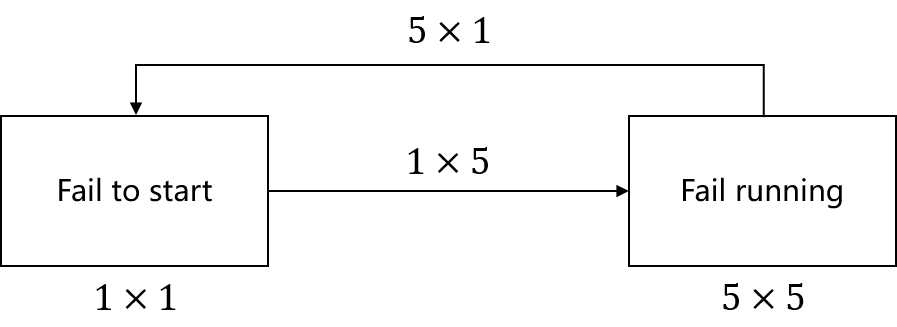}
\caption{The relationships among various variables needed to be consulted.}\label{influence}
\end{figure}

(3.1) Establish pair-wise influential relationships among BN variables from domain experts. To establish causal relationships among the 6 variables in Table \ref{failure_mode}, we consulted three experts (recruited from the nuclear science and technology professionals) through two rounds of questionnaire surveys based on the Delphi method \cite{linstone1975delphi}. In the first round, experts are asked to judge on the relationship between each pair of variables (For example, “Does the SI-P2-DF ($N_{1}$) influences the SI-P1-RF ($N_{2}$)?”). If more than two experts reach the consensus about a relationship, it will be determined. Otherwise, it would be ‘undecided.’ In the second round, we summarized all experts’ opinion about the relationships and sent the summary to the experts who made re-assessment on those “undecided” relationships until reaching the consensus. The size of the questionnaire is 36 relationships shown in Figure \ref{influence}. 

(3.2) Establish the Structural Self-Interaction Matrix SSIM in Equation\ref{SSIM}. 

\begin{equation}
SSIM = \left [ s_{ij}  \right ] _{n\times n} 
\label{SSIM}
\end{equation}

For any pair of variables $N_i$ and $N_j$ from the set $N={N_{1},...,N_{n} }$, $s_{ij}$ denotes the direct relationship between them. which is illustrated in Equation \ref{s_{ij}}.
\begin{equation}
s_{ij} = \begin{cases}
1, & \text{if } N_i \text{ is related to } N_j \\
0, & \text{if there is no relationship between } N_i \text{ and } N_j
\end{cases}
\label{s_{ij}}
\end{equation}

Based on expert survey, a SSIM is derived:

\begin{figure}[H]
\centering
\includegraphics[width=0.6 \textwidth]{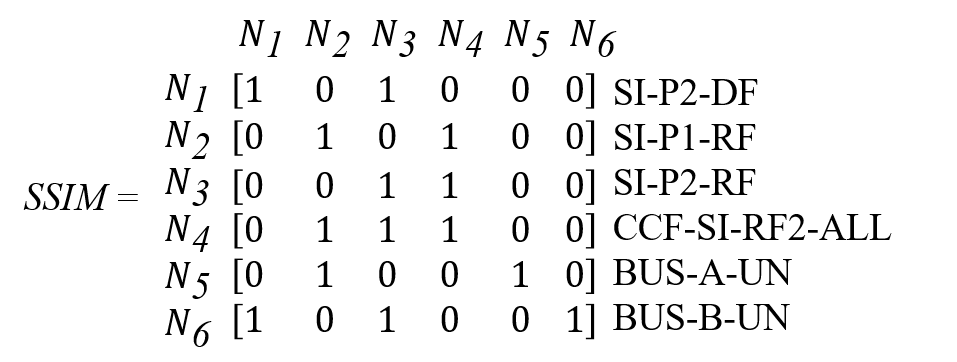}
\caption{Simplified system diagram for a NPP}\label{SSIM}
\end{figure}

4. Calculate the reachability matrix M from SSIM. The reachability matrix (M) denotes whether there is a connected path from one variable to another. If node $N_{i}$ can be connected to node  $N_{j}$ after a certain length of path, it is considered reachable. If $N_{i}$ needs to go through $l$ branches to reach $N_{j}$, the length from $N_{i}$ to $N_{j}$ is $l$. The matrix M can be expressed as a binary $n_{i}$ × $n_{j}$ matrix, and each element $m_{ij}$ of the matrix can be expressed as Equation \ref{m_{ij}}.

\begin{equation}
m_{ij} =
\begin{cases} 
1 & \text{if } S_i \text{ to } S_j \text{ is reachable} \\
0 & \text{if } S_i \text{ to } S_j \text{ is unreachable} 
\end{cases}
\label{m_{ij}}
\end{equation}

\begin{equation}
M = SSIM \lor SSIM^2 \lor SSIM^3 \lor \cdots \lor SSIM^{n-1} \lor SSIM^n
\label{M}
\end{equation}

\begin{figure}[H]
\centering
\includegraphics[width=0.6 \textwidth]{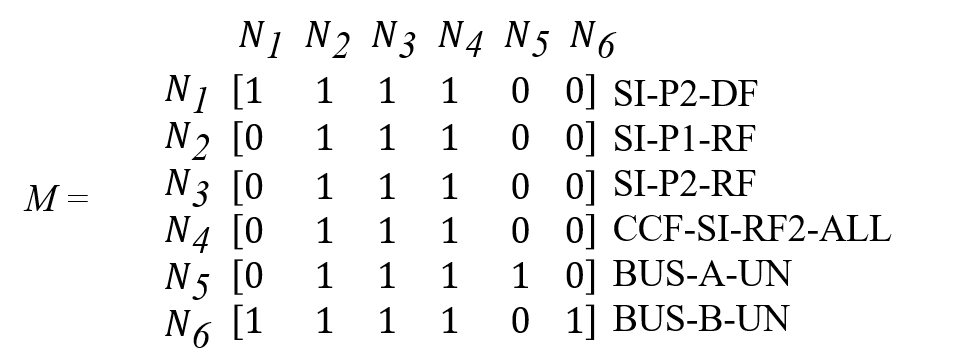}
\caption{ The reachability matrix (M)}\label{M}
\end{figure}

5. Partition M into different levels to construct the influence diagram. Define the reachability set $R(X_{i})$ as elements that are reachable from $X_{i}$ (including $X_{i}$), and the antecedent set $A(X_{i})$ as the elements that can reach $X_{i}$ (including $X_{i}$). We can derive a level for each element according to $C(N_{i}) = R(N_{i})\cap A(N_{i})$. Once the top-level elements are identified, they will be separated out from the other elements. The same process undergoes iterations until the levels of all elements are achieved.  

\begin{table}[H] 
\centering 
\begin{tabular}{p{1cm} p{4cm} p{5cm} p{2cm}} 
\hline
Item & Reachability set & Antecedent set  & Intersection set \\ 
\hline
$N_{1}$ &  $N_{1}$, $N_{2}$, $N_{3}$, $N_{4}$  & $N_{1}$, $N_{6}$   &$N_{1}$ \\
$N_{2}$ & $N_{2}$, $N_{3}$, $N_{4}$ & $N_{1}$, $N_{2}$, $N_{3}$,  $N_{4}$, $N_{5}$, $N_{6}$  & $N_{2}$, $N_{3}$, $N_{4}$ \\
$N_{3}$ & $N_{2}$, $N_{3}$, $N_{4}$ & $N_{1}$, $N_{2}$, $N_{3}$, $N_{4}$, $N_{5}$, $N_{6}$ & $N_{2}$, $N_{3}$, $N_{4}$ \\
$N_{4}$ & $N_{2}$,$N_{3}$,$N_{4}$ & $N_{1}$, $N_{2}$, $N_{3}$,  $N_{4}$, $N_{5}$, $N_{6}$  & $N_{2}$, $N_{3}$, $N_{4}$ \\
$N_{5}$ & $N_{2}$, $N_{3}$, $N_{4}$, $N_{5}$ & $N_{5}$ & $N_{5}$ \\
$N_{6}$ & $N_{1}$, $N_{2}$, $N_{3}$, $N_{4}$, $N_{6}$  & $N_{6}$ & $N_{6}$ \\
\hline
\end{tabular}
\caption{The Parameters  $\alpha$,  $\beta$, and $\gamma$ in the HCR Model}
\label{HEPs}
\end{table}

When the levels of all factors are determined by the above process, the variables are divided into 4 levels:

\begin{itemize}
    \item Level 1:  BUS-B-UN ($N_{6}$)
    \item Level 2:  BUS-A-UN ($N_{5}$)
    \item Level 3:  SI-P1-RF ($N_{2}$), SI-P2-RF($N_{3}$), CCF-SI-RF2-ALL ($N_{4}$)
    \item Level 4:  SI-P2-DF ($N_{1}$)
\end{itemize}

Then, establish the direct connections and derive the corresponding directed acyclic graph (DAG). The hierarchical structure and interrelationships between the basic events are depicted in Figure \ref{levels} (a), which represents the structure of the virtual fault tree. Similarly, the virtual fault tree for the CS system is illustrated in Figure \ref{levels} (b).

\begin{figure}[H]
\centering
\includegraphics[width=1.0 \textwidth]{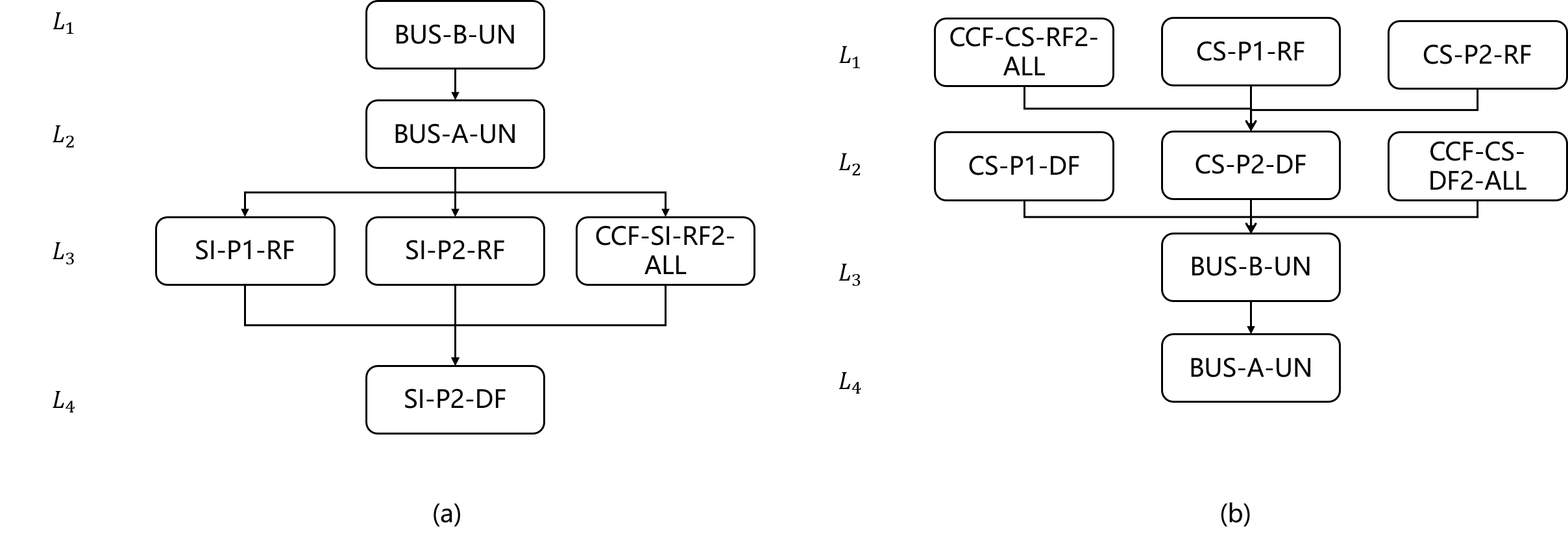}
\caption{The hierarchical structure of safety injection  and containment spray systems}\label{levels}
\end{figure}


\subsection{Implementation Details}

Our model is based on a Graph Convolutional Network (GCN) and consists of three graph convolutional layers. The learning rate is set to 0.001, which controls the speed of weight updates in the model. The feature dimension of the hidden layers is set to 32, while each node has only one feature, representing the probability of the basic event occurring.

\subsection{Evaluation Metrics}

As illustrated in Section \ref{Performance Evaluation and Validation},MSE (Mean Squared Error), RMSE (Root Mean Squared Error), MAE (Mean Absolute Error), and $R^{2}$  (R-squared, coefficient of determination) are commonly used evaluation metrics in machine learning and statistical models for regression. The specific definitions of these metrics are following.

MSE is the average of the squared differences between the predicted and actual values, used to measure the overall error of the model's predictions. It is more sensitive to larger errors, as the squaring of the error increases the impact of larger deviations. The formula for calculating MSE is shown in Equation \ref{MSE}.

\begin{equation}
MSE = \frac{1}{n} \sum_{n}^{i=1}(y_{i}-\hat{y}_{i})^{2}   
\label{MSE}
\end{equation}
where, $y_{i}$ is the actual value, $\hat{y}_{i}$ is the predicted value,$n$ is the number of samples.

RMSE is the square root of MSE, retaining the same units as the actual data, making it easier to interpret. It is often used to measure the extent of variation in predictions and their proximity to actual values. The formula for calculating RMSE is shown in Equation \ref{RMSE}.

\begin{equation}
RMSE =\sqrt{MSE} = \sqrt{\frac{1}{n} \sum_{n}^{i=1}(y_{i}-\hat{y}_{i})^{2}}      
\label{RMSE}
\end{equation}

MAE is the average of the absolute differences between the predicted and actual values. Unlike MSE, it does not amplify larger errors, making it less sensitive to outliers. It provides the average magnitude of the prediction errors. The formula for calculating MAE is shown in Equation \ref{MAE}.
\begin{equation}
MAE = \frac{1}{n} \sum_{i=1}^{n}\left | y_{i} - \hat{y_{i}}   \right |  
\label{MAE}
\end{equation}

$R^{2}$ measures the goodness of fit of a model, indicating the proportion of variance in the dependent variable that is explained by the model. The value of $R^{2}$ ranges from 0 to 1, with values closer to 1 indicating better model fit. A negative $R^{2}$ value suggests that the model performs worse than a simple mean prediction. The formula for calculating $R^{2}$ is shown in Equation \ref{R²}.
\begin{equation}
R^{2} =1- \frac{ \sum_{i=1}^{n}(y_{i} - \hat{y_{i}}) }{\sum_{i=1}^{n} (y_{i} - \bar{y})}  
\label{R²}
\end{equation}

\section{Results and Analysis}
\label{Results and Analysis}

\subsection{Comparative Network Analysis}

To further evaluate the effectiveness of our virtual fault tree, we compared the structure of our ISM with a network structure generated using a data-driven approach through HillClimbSearch \cite{lalouni2015maximum}. HillClimbSearch is a greedy algorithm designed to iteratively adjust the network structure in search of one that maximizes a given scoring function. For this purpose, we selected the Bayesian Dirichlet Equivalent Uniform (BDeu) score \cite{scutari2016empirical}, a scoring function particularly suited for handling discrete data. The BDeu score assesses the network structure by incorporating Bayesian statistics and prior information, allowing for a robust evaluation of the generated network. The structural networks generated by the CS and SI systems are illustrated in Figure 9, specifically in subfigure (a) and subfigure (b), respectively.

\begin{figure}[H]
\centering
\includegraphics[width=1.0 \textwidth]{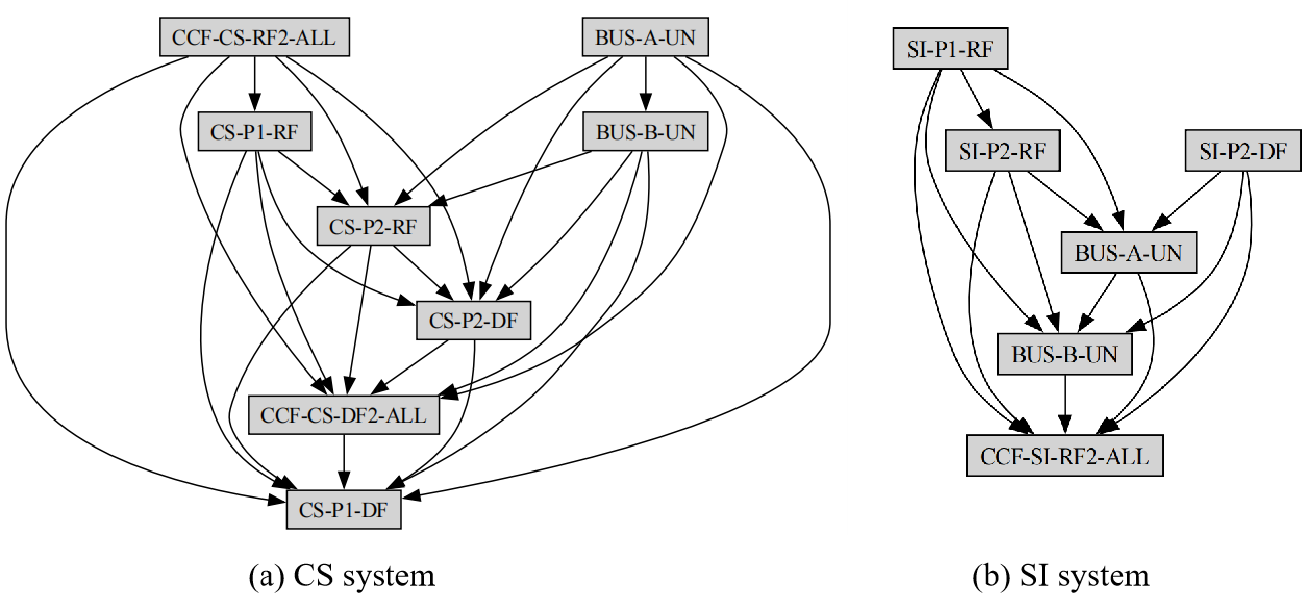}
\caption{ Structural representation of basic event dependencies identified via HillClimbSearch}\label{hill}
\end{figure}

The table \ref{Structures_compare} demonstrates that the ISM-based structure significantly outperforms the HillClimbSearch-derived structure. Specifically, the ISM-based structure achieves a 32.69\% improvement in MSE (0.0035 vs. 0.0052), an 18.10\% improvement in RMSE (0.0588 vs. 0.0718), and a 43.13\% improvement in MAE (0.0236 vs. 0.0415). Additionally, the ISM-based structure shows a 1.08\% improvement in $R^{2}$ (0.9788 vs. 0.9684). These results highlight the significant performance gains offered by the ISM-based structure over the HillClimbSearch-derived structure.

\begin{table}[H] 
\centering 
\begin{tabular}{p{4cm} p{2cm}p{2cm}p{2cm}p{2cm} } 
\hline
\centering Model & MSE & RMSE & MAE & $R^{2}$\\ 
\hline
 \centering ISM & 0.0035 & 0.0588 & 0.0236& 0.9788 \\ 
\centering HillClimbSearch & 0.0052 & 0.0718 & 0.0415&0.9684 \\
\centering Improvement & 32.69\% & 18.10\% & 43.13\% &1.08\% \\
\hline
\end{tabular}
\caption{Performance Comparison of ISM-Derived and HillClimbSearch-Derived Structures}
\label{Structures_compare}
\end{table}

\subsection{ Model Performance Evaluation}

The performance of our model is shown in Table \ref{results}. The MSE and RMSE are 0.0191 and 0.1381, respectively, both of which are relatively low, indicating that the model has small prediction errors and performs well on the test set. The MAE is 0.0979, also low, meaning the errors are within an acceptable range, demonstrating that the model's predictions are stable without significant deviations. Additionally, the $R^{2}$ is 0.8832, suggesting that the model is able to explain a large portion of the variance in the data, indicating a good fit.

To evaluate the performance of our model compared to other models, we implemented a Multi-Layer Perceptron (MLP) for comparison. To ensure consistency, the MLP was configured with 4 layers and a learning rate of 0.001. We input the probabilities of fundamental events along with their logical relationships, while the output was the feature vector (FV) importance for each fundamental event.  

The core equation of the MLP is given by equation \ref{MLP_equation}, where: $a_{j}$ represents the output of the neuron, $\sigma$ denotes the activation function, $w_{ij}$ is the weight between the $i$-th input and the $j$-th output, $x_{i}$ refers to the $i$-th input, $b_{j}$ is the bias for the $j$-th output, and $n$ represents the number of inputs. 
\begin{equation}
a_{j} =\sigma (\sum_{i=1}^{n}w_{ij}x_{i} +b_{j}   )
\label{MLP_equation}
\end{equation}

The results, shown in Table \ref{results}, demonstrate that our model slightly outperforms the MLP.

\begin{figure}[H]
\centering
\includegraphics[width=0.7 \textwidth]{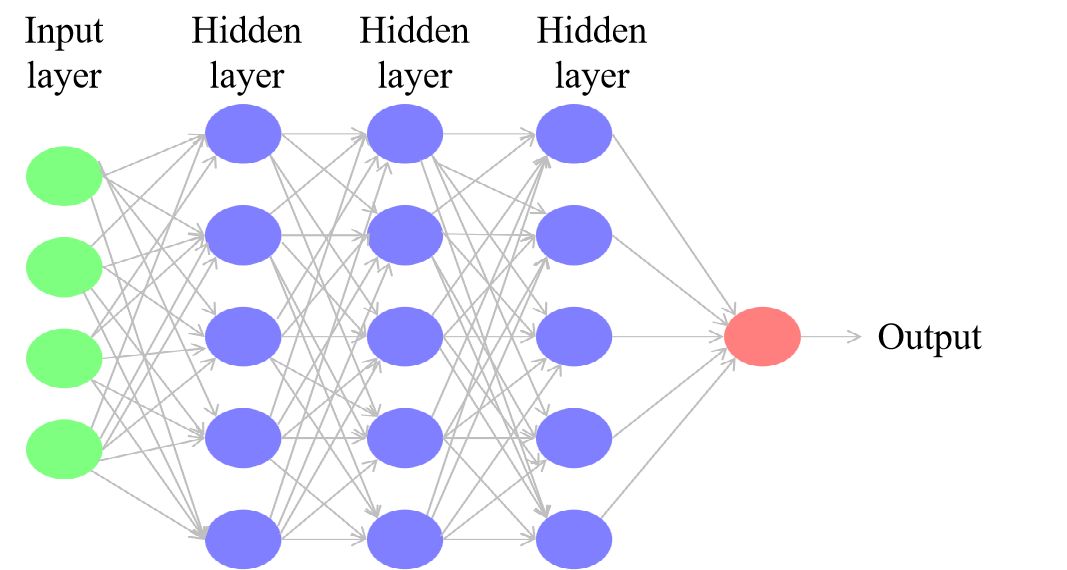}
\caption{Architecture of the constructed four-layer MLP for comparative experiments.}\label{prompt}
\end{figure}

To evaluate the effectiveness of large language models for this task, we selected Claude 3.5 Sonnet for testing. Claude 3.5 Sonnet supports a 200K-token context window and has shown exceptional performance in multiple evaluations, outperforming both OpenAI's leading model, GPT-4o, and its predecessor, Claude 3 Opus \cite{kurokawa2024diagnostic}. With the assistance of specific instructions and tools, Claude 3.5 Sonnet can autonomously write, edit, and execute code, while also demonstrating advanced reasoning and problem-solving skills. Additionally, it maintains the speed and cost efficiency commonly associated with mid-sized models \cite{bae2024enhancing}.

\begin{figure}[H]
\centering
\includegraphics[width=1.0 \textwidth]{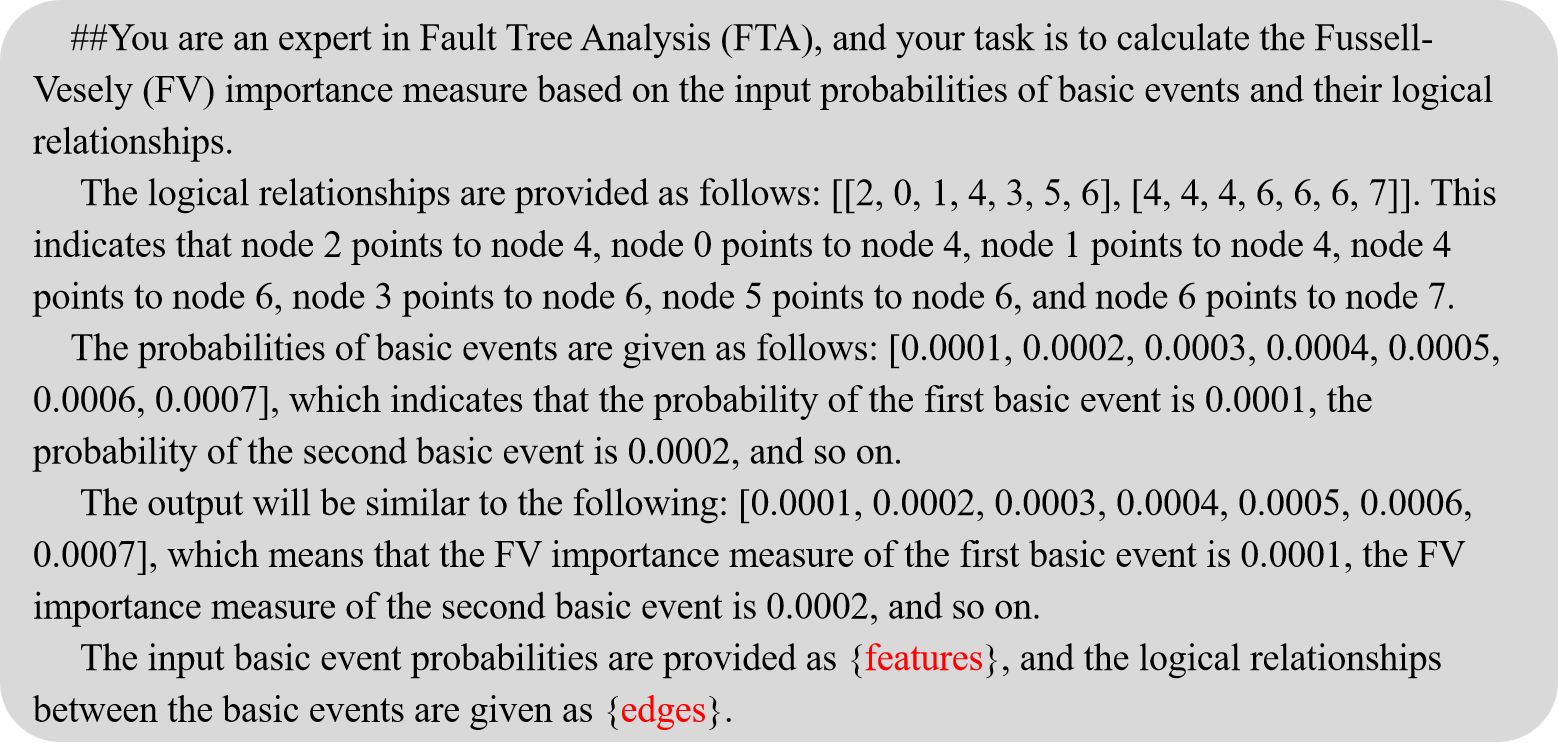}
\caption{ The prompt content with Claude 3.5 Sonnet for solving FV importance}\label{prompt}
\end{figure}

Prompting is a simple yet effective method for ensuring large language models adhere to specific instructions, and it has been widely utilized across numerous tasks \cite{zamfirescu2023johnny}. To facilitate Claude 3.5 Sonnet in performing the FV importance analysis, we developed the prompt as shown in Figure \ref{prompt}. In the Figure \ref{prompt}, the red features represent the occurrence probabilities of basic events, while the edges depict the constructed virtual fault tree relationships. Unfortunately, the performance metrics of Claude 3.5 Sonnet were far from satisfactory. The model exhibited a MSE of 2,333,439,021.98, a MAE of 2,986.07, and an exceedingly high RMSE of 48,305.68, all of which indicate a considerable deviation between predicted and actual values. Furthermore, the coefficient of $R^{2}$ was calculated to be -14,284,500,759.08, revealing that the model performed even worse than a simple mean-based predictor, entirely failing to extract meaningful patterns from the data.

\begin{table}[H] 
\centering 
\begin{tabular}{p{2cm} p{2cm}p{2cm}p{4cm} } 
\hline
\centering Metric & Our Model & MLP & Claude 3.5 \\ 
\hline
 \centering MSE & 0.0035 & 0.0036 & 2,333,439,021.9810\\
\centering RMSE & 0.0588 & 0.0589& 2,986.0738\\
\centering MAE & 0.0236 &0.0259 & 48,305.6831\\
\centering $R^{2}$ & 0.9788 & 0.9787& -14,284,500,759.0832\\
\hline
\end{tabular}
\caption{Performance metrics comparison of our Model, MLP, and Claude: MSE, RMSE, MAE, and $R^{2}$ values}
\label{results}
\end{table}

\subsection{ Time Efficiency Comparison}

Furthermore, to demonstrate the computational efficiency of our framework, we conducted an evaluation involving a PhD student specialized in the field. The test focused on solving the Fault Tree Analysis (FTA) of FV importance for two systems. We first measured the time required to construct both a traditional fault tree and a virtual fault tree. The results showed that the construction times were 1 minutes and 18.70 seconds and 2 minutes and 0.16 seconds, respectively (time units to be clarified). This indicates that our virtual fault tree offers a significant advantage in terms of construction efficiency. 

\begin{figure}[H]
\centering
\includegraphics[width=1.0 \textwidth]{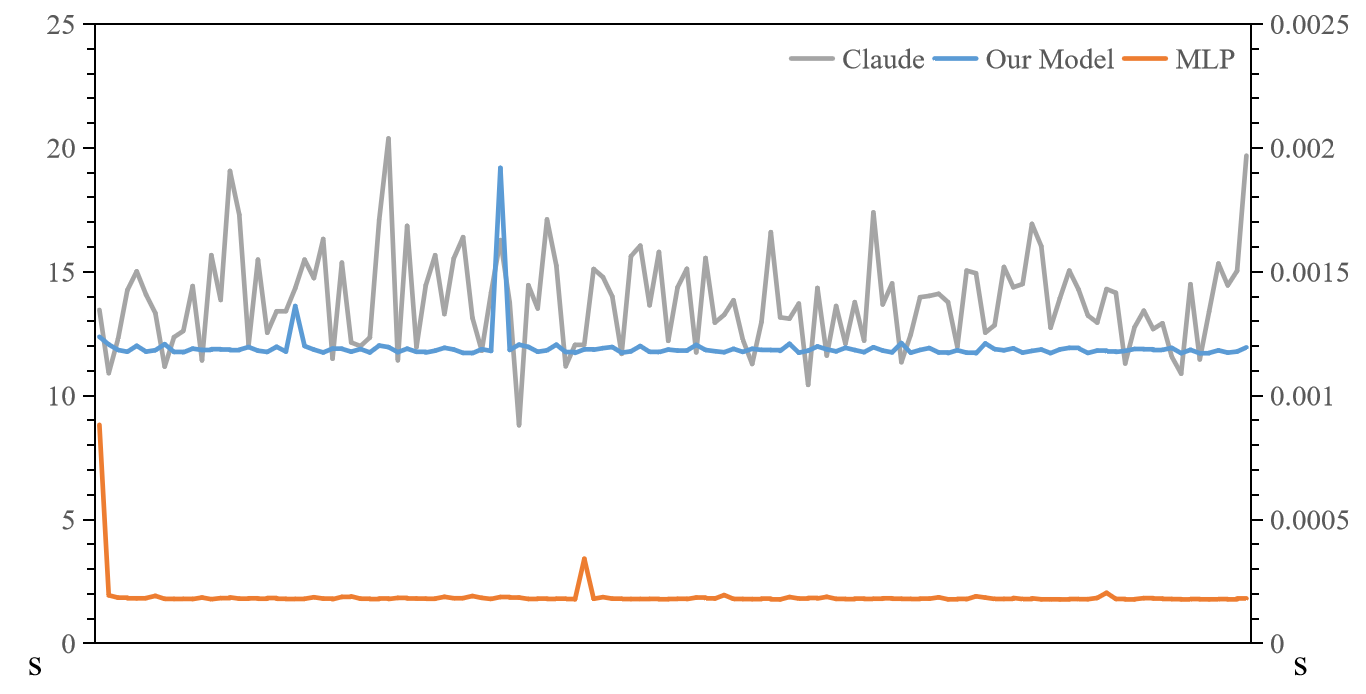}
\caption{Comparison of inference time across GNN, MLP, and Claude algorithms}\label{time}
\end{figure}

Subsequently, we compared the time required to calculate FV importance manually with that of the model-based computations. The results indicate that manual calculations took 2 minutes and 5.19 seconds, whereas the MLP and GNN models performed the calculations within milliseconds in Figure \ref{time}. Additionally, the computation time for the Claude is relatively higher but remains within 25 seconds. This clearly demonstrates the superiority of our proposed method.

\subsection{Case study: Scenario Deduction and Decision Support}

Our data only includes two types of structures, which limits the model's ability to generalize to system structures.  Taking the SI system as an example, the probabilities of the basic events are shown in the first column of the Table~\ref{case_study}, our model's predicted FV importances are in the second column, and the actual FV importances are in the third column. Since this is a regression task, there are some differences in the specific numerical values.

However, our results indicate that the feature importance ranking generated by our model aligns well with the actual observations. Specifically, the events SI-P2-DF and SI-P1-RF contribute significantly to the model, whereas the relative importance of SI-P2-RF and CCF-SI-RF2-ALL is considerably lower. The contributions of BUS-A-UN and BUS-B-UN are negligible. Thus, our model can provide a real-time ranking of the basic events in this scenario and offer key information for controlling these events to achieve effective risk management.

\begin{table}[H] 
\centering 
\begin{tabular}{p{3.5cm} p{2cm}p{3.2 cm}p{3.2 cm} } 
\hline
\centering Basic Event & Probability & Predicted Results & Reference\\ 
\hline
 \centering SI-P2-DF &9.9e-01 & 9.9497789e-01 & 1.0000 \\
\centering SI-P1-RF & 9.9e-01 &9.9497789e-01&1.0000\\
\centering SI-P2-RF & 1.0e-02 &7.4510276e-03 &0.0102\\
\centering CCF-SI-RF2-ALL & 2.0e-05&  1.9852519e-03&2.02e-05\\
\centering BUS-A-UN & 5.5e-07 &2.9288232e-04&5.55e-07\\
\centering BUS-B-UN & 5.5e-07& 2.9288232e-04&5.55e-07\\

\hline
\end{tabular}
\caption{The results of our framework}
\label{case_study}
\end{table}

\section{Conclusion and Discussion}
\label{Conclusion and Discussion}

This study presents a novel hybrid real-time framework for evaluating Fussell-Vesely (FV) importance in complex systems, effectively addressing several key challenges faced by traditional reliability analysis methods. By integrating Interpretive Structural Modeling (ISM) with Graph Neural Networks (GNN), our framework bridges the gap between expert-driven structural modeling and data-driven learning, offering a more efficient and dynamic approach to system reliability assessment. The main contributions are as follows.

\begin{itemize}
\item The primary contribution of this research lies in the development of a virtual fault tree that exclusively focuses on basic events, reducing the complexity typically associated with traditional fault tree analysis. Through ISM, expert knowledge is systematically incorporated to capture the dependencies between events, providing a clear and simplified hierarchical structure without the need to account for intermediate events. This structural reduction enhances the manageability of reliability models, making them less cumbersome and more adaptable to real-world applications.

\item Another significant contribution of this study is the introduction of GNN as a tool for real-time FV importance evaluation. By feeding the ISM-based virtual fault tree into a GNN, the framework leverages real-time system data to continuously update and recalibrate FV importance scores. This integration not only reduces the time required to perform reliability analysis but also enables dynamic, real-time decision support, a critical advantage in rapidly changing environments. The GNN’s capacity to capture complex dependencies between basic events ensures a more accurate and responsive risk assessment, even in the presence of evolving system conditions.
    
\end{itemize}

In conclusion, this study contributes a robust and innovative framework that effectively balances the strengths of expert knowledge and machine learning to provide more efficient, accurate, and dynamic reliability assessments. The hybrid approach opens new pathways for enhancing the evaluation of FV importance, offering significant improvements in both computational efficiency and decision-making capability, especially in complex, rapidly evolving systems. However, due to data limitations, we only conducted tests on small-scale fault trees. Future work could explore expanding the types of data and discussing the applicability of our framework to large-scale fault trees.

\section{Acknowledgements}

The research was supported by the Innovation Funds of CNNC–Tsinghua Joint Center for Nuclear Energy R\&D (Project No. 20202009032) and a grant from the National Natural Science Foundation of China (Grant No. T2192933).

\section{Author contributions statement}
Xiao Xingyu: Methodology, Software, Formal analysis, Data Curation, Visualization,Validation, Writing- Original draft preparation. Peng Chen: Software, Methodology. Qi Ben: Software, Methodology, Writing - Review and Editing. Peng Pengcheng: Data Curation. Liang Jingang: Conceptualization, Resources, Supervision, Writing - Review and Editing, Project administration, Funding acquisition. Tong Jiejuan: Investigation, Supervision, Writing - Review and Editing. Wang Haitao: Supervision, Writing- Reviewing and Editing.

\section{reference}

\end{document}